\documentclass{bmvc2k}
\usepackage{subcaption}
\usepackage{caption}
\usepackage{graphicx}

\usepackage{amsmath}
\usepackage{amssymb}
\usepackage{booktabs}
\usepackage{multirow}
\usepackage{bbm}
\usepackage{array}
\usepackage{makecell}

\title{Exploring Non-additive Randomness on ViT against Query-Based Black-Box Attacks}

\addauthor{Jindong Gu}{jindong.gu@outlook.com}{1}
\addauthor{Fangyun Wei}{fawe@microsoft.com}{2}
\addauthor{Philip Torr}{philip.torr@eng.ox.ac.uk}{1}
\addauthor{Han Hu}{hanhu@microsoft.com}{2}

\addinstitution{
 Torr Vision Group\\
 Department of Engineering Science\\
 University of Oxford\\
 Oxford, UK
}
\addinstitution{
 Microsoft Research Asia\\
 Beijing, China
}

\runninghead{Jindong \etal}{Non-additive Randomness of ViT against QBBA}

\def\etal{\emph{et al}\bmvaOneDot}

\begin{document}

\maketitle

\begin{abstract}
Deep Neural Networks can be easily fooled by small and imperceptible perturbations. The query-based black-box attack (QBBA) is able to create the perturbations using model output probabilities of image queries requiring no access to the underlying models. QBBA poses realistic threats to real-world applications. Recently, various types of robustness have been explored to defend against QBBA. In this work, we first taxonomize the stochastic defense strategies against QBBA. Following our taxonomy, we propose to explore non-additive randomness in models to defend against QBBA. Specifically, we focus on underexplored Vision Transformers based on their flexible architectures. Extensive experiments show that the proposed defense approach achieves effective defense, without much sacrifice in performance.
\end{abstract}

\section{Introduction}
The decisions of deep neural networks (DNNs) can be misled by imperceptible perturbations. The adversary perturbations can be created only using the model output scores (i.e., softmax probabilities) of image queries and require no access to the underlying models~\cite{chen2017zoo,brendel2017decision}. As a result, DNN-based real-world applications face the potential threats posed by query-based black-box attacks (QBBA).

Threat Model of QBBA: For this type of attack, attackers can only access model output scores, while they cannot get access to the model architecture, model parameters, input gradients, or defense strategies. Many efficient and effective such attacks have raised the attention of the community~\cite{cheng2018query,bhagoji2018practical,andriushchenko2020square}. Recent work explores various types of randomness to defend against QBBA~\cite{xie2017mitigating,he2019parametric,byun2022effectiveness,qin2021random}. For example, the randomness in inputs or models can mislead the gradient estimation or search process of QBBA. 

To articulate different types of defensive randomness,
in this work, we first taxonomize the defense strategies against QBBA. Following our taxonomy, we propose to explore non-additive randomness in models to defend against QBBA. Specifically, we focus on Vision Transformers (ViT)~\cite{dosovitskiy2020image} models for the following reasons: 1) As an alternative to CNNs, ViTs have received great attention; 2) The non-additive randomness on ViT is underexplored; 3) The network architecture is flexible.

Different from CNNs, ViT represents an input image as a sequence of image patches and learns image representations by applying the self-attention module to the patches. We propose three ways to integrate non-additive robustness in the self-attention module of ViT and provide analyses of their effectiveness against QBBAs. We verify our approach on popular QBBAs and show that ours can defend against them without sacrificing clean performance.

\section{Related Work}
\noindent\textbf{Query-based Black-box Attacks.} The popular black-box attacks include transfer-based black-box attacks and quey-based black-box attacks (QBBA). The former leverages the observation that adversary examples created on one model can be transferred to different model architectures, e.g., Vision Transformers~\cite{benz2021adversarial,shi2021decision}, Capsule Networks~\cite{gu2021effective}, Spiking Neural Networks~\cite{yu2023reliable}. QBBA can be categorized into score-based attacks and decision-based ones according to the availability of model outputs. The score-based attacks are able to access class-wise output probabilities~\cite{chen2017zoo,bhagoji2018practical,tu2019autozoom,ilyas2018black,alzantot2019genattack,guo2019simple,uesato2018adversarial,al2019there,meunier2019yet,moon2019parsimonious,shukla2019black,andriushchenko2020square}, while decision-based ones assume only the top-1 class index is available~\cite{brendel2017decision,cheng2018query,cheng2019sign,liu2019geometry,chen2020hopskipjumpattack,rahmati2020geoda}. On the other hand, according to the method to create adversarial examples, query-based
attacks can be also categorized into optimization-based attacks and search-based attacks. The former estimates gradients of the loss with respect to the input to optimize a pre-defined adversary’s objective loss~\cite{chen2017zoo,bhagoji2018practical,ilyas2018black,uesato2018adversarial,shukla2019black,cheng2018query,cheng2019sign}. In contrast, the latter repeatedly searches for the desired images from randomly perturbed ones so that the desired ones get closer to adversary’s objective~\cite{al2019there,alzantot2019genattack,guo2019simple,meunier2019yet,moon2019parsimonious,andriushchenko2020square,andriushchenko2020square,brendel2017decision,liu2019geometry,chen2020hopskipjumpattack,rahmati2020geoda}.

\vspace{0.1cm}
\noindent\textbf{Defense against Query-based Black-box Attacks.}
Most defense strategies are designed for defending against white-box attacks. For example, adversarial training~\cite{madry2017towards}. The adversarially trained models also show certain robustness against QBBA, but often with a sacrifice in clean performance~\cite{madry2017towards,xie2020adversarial}. 
Randomness has also been intensively explored to defend against both white-box and black-box attacks. The works~\cite{liu2018towards,he2019parametric} propose to add Gaussian noise to model parameters and intermediate activations to boost model robustness. Random transformation of input, as a pre-process defense strategy is proposed in the work~\cite{xie2017mitigating}. The works~\cite{byun2022effectiveness,qin2021random} show that simply adding small noise to inputs can effectively defend against black-box attacks. The relationship between these stochastic defense strategies remains to be articulated.

\vspace{0.1cm}
\noindent\textbf{Adversarial Robustness of ViT.} The robustness of ViT have achieved great attention due to its great success in many vision tasks~\cite{bhojanapalli2021understanding,shao2021adversarial,paul2021vision,naseer2021intriguing,bai2021are,benz2021adversarial,shi2021decision,pinto2022impartial,gu2022vision}. The works~\cite{wu2022towards,bhojanapalli2021understanding,shao2021adversarial,benz2021adversarial} demonstrate that ViT achieves higher adversarial robustness than CNNs under white adversarial attacks. The work~\cite{pinto2022impartial,gu2022vision} shows a different observation in a test setting, e.g., under distribution shift and patch attack. The work~\cite{benz2021adversarial,shi2021decision} makes the exploration of QBBA attacks and transfer-based attacks on ViTs. They show ViT also suffers from the basic QBBA, such as boundary attacks and Bandits-based attacks. The works~\cite{wu2022towards,shao2021adversarial,tang2021robustart} show adversarial training can also be applied to boot the adversarial robustness of ViT. The simple patch permutation of ViT has also been explored to study its robustness~\cite{naseer2021intriguing,shao2021adversarial,doan2022defending,gu2022vision}. In this work, we systematically explore non-additive randomness of ViT against QBBA. \vspace{-0.5cm}

\section{Stochastic Defense of ViT against Query-Based Black-box Attacks}
In this work, we first present our taxonomy the stochastic defense strategies. Following our taxonomy, we propose non-additive randomness on ViT to defend against QBBA. Then, we show our method performs better than the rest of the methods in our taxonomy.  \vspace{-0.3cm}

\subsection{Taxonomy of Stochastic Defense Strategy}
\begin{figure}[t]
    \centering
    \includegraphics[scale=0.45]{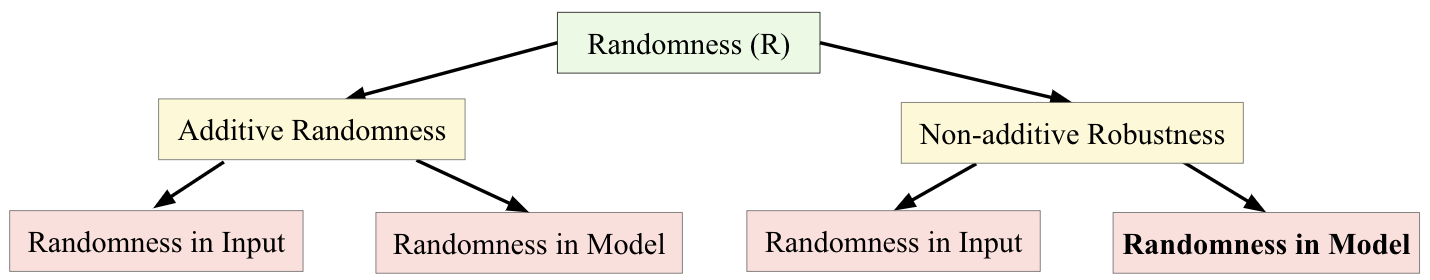} \vspace{-0.15cm}

    {\hspace{-0.5cm}\footnotesize SND\cite{byun2022effectiveness,qin2021random} \hspace{1.0cm} RSE\cite{liu2018towards}, PNI\cite{he2019parametric}  \hspace{1.3cm}   R\&P\cite{xie2017mitigating} \hspace{1.9cm}  Ours}
    \caption{Taxonomy of Stochastic Defense Strategy. The randomness can be categorized into additive randomness and non-additive one. Each of the two types randomness can be integrated either into the inputs or into the model.}
    \label{fig:rand_overview}
\end{figure}

Randomness has been explored intensively to defend against attacks. The earlier exploration in~\cite{liu2018towards} adds trainable gaussian noise into intermediate activations to build a robust model, which was shown to be robust to white-box attacks. A later work~\cite{xie2017mitigating} shows the random transformation of input can be used to defend against both white-box and black-box attacks. The recent work~\cite{byun2022effectiveness,qin2021random} shows simply adding small noise to inputs is surprisingly effective to defend against QBBA. Without a doubt, all these methods can be applied to defend against QBBA. In this work, we attempt to articulate the relationship among them and taxonomize the existing stochastic defense strategies, as shown in Fig.~\ref{fig:rand_overview}.

In our taxonomy, we categorize randomness into two types, namely, additive randomness and non-additive one. The additive randomness is often implemented by adding small noise to the input or the model parameters. In contrast, the non-additive randomness integrated non-additive perturbation into the input or the model. For example, the affine transformation of inputs is one type of non-additive perturbation. As shown in taxonomy, the non-additive randomness on the model has not been explored yet to defend against QBBA. In the next subsection, we describe our proposal to fill the vacancy.

\subsection{Non-additive Randomness in ViT for Defense}
We first review the background knowledge of ViT and then present our three ways to integrate non-additive randomness into ViT model.

\vspace{0.05cm}
\noindent\textbf{Background knowledge on ViT} Different from CNNs, ViT represents an input image as a sequence of image patches. Then, a list of self-attention modules is applied to the sequence of image patches sequentially. We now introduce the details of the primary ViT architecture in~\cite{dosovitskiy2020image}. The input image $\boldsymbol{X} \in \mathbb{R}^{(H \times W \times C )}$ is first split into image patches $\{\boldsymbol{x}_i\in \mathbb{R}^{P\times P\times C}\vert i \in (1,2,3, ..., {HW/P^2}) \}$ where $P$ is the patch size. 

The embedding of each patch is extracted from the raw image patch with linear projection. Before the application of self-attention module, the position information of image patches is also integrated into the patch embedding. The embedding of the image patches in the input layer is denoted as $\{\boldsymbol{Z}^0_i\in \mathbb{R}^{D}\vert i \in (1,2,3, ..., {HW/P^2}) \}$. A learnable class-token embedding $\boldsymbol{Z}^0_0$ is added into the list of patch embeddings. The class embedding in the last layer is taken as the image embedding for classification.

We now introduce the transformer encoder where the list of blocks is applied to transform the input embeddings. Each block consists of two main modules, namely, a multi-head self-attention module $MSA(\cdot)$ to model the inter-patch relationship and an MLP module $MLP(\cdot)$ to project patch embeddings into a new space, respectively.

When the self-attention module in $(l+1)$-th layer is applied to input patches $\{\boldsymbol{Z}^l_i\in \mathbb{R}^{D}\vert i \in (0,1,2, ..., {HW/P^2})\}$ in the $l$-th layer, the attention of $i$-th patch in a single head is computed as follows
$
\boldsymbol{K}^{l+1}_i = \boldsymbol{W}^{l+1}_k \cdot \boldsymbol{Z}^l_i,  \;\;\;\; 
\boldsymbol{Q}^{l+1}_i = \boldsymbol{W}^{l+1}_q \cdot \boldsymbol{Z}^l_i, \;\;\;\;
\boldsymbol{A}^{l+1}_i =  Softmax(Q^{l+1}_i(K^{l+1})^T/\sqrt{D}),
$
The output embedding of the $i$-th patch $\boldsymbol{Z}^l_i$ is  
$
\boldsymbol{V}^{l+1}_i = \boldsymbol{W}^{l+1}_v \cdot \boldsymbol{Z}^l_i, \;\;\;\;\;
\boldsymbol{Z}^{l+1}_i = \sum^{{HW/P^2}}_{j=0} \boldsymbol{A}^{l+1}_{ij} \cdot V^{l+1}_j.
\label{equ:patch_val}
$

Then, an MLP module with two MLP layers is applied to project the final embedding of each patch into a new feature space. The embeddings of $i$-th patch in different heads are concatenated as its final embedding in the $(l+1)$-th layer.  The embedding of the class-token patch in the last block is taken as the image representation to classify the image. A linear classifier maps the features to output space.

\begin{figure}[!t]
\centering
    \begin{subfigure}[b]{0.3\textwidth}
    \centering
    \includegraphics[scale=0.15]{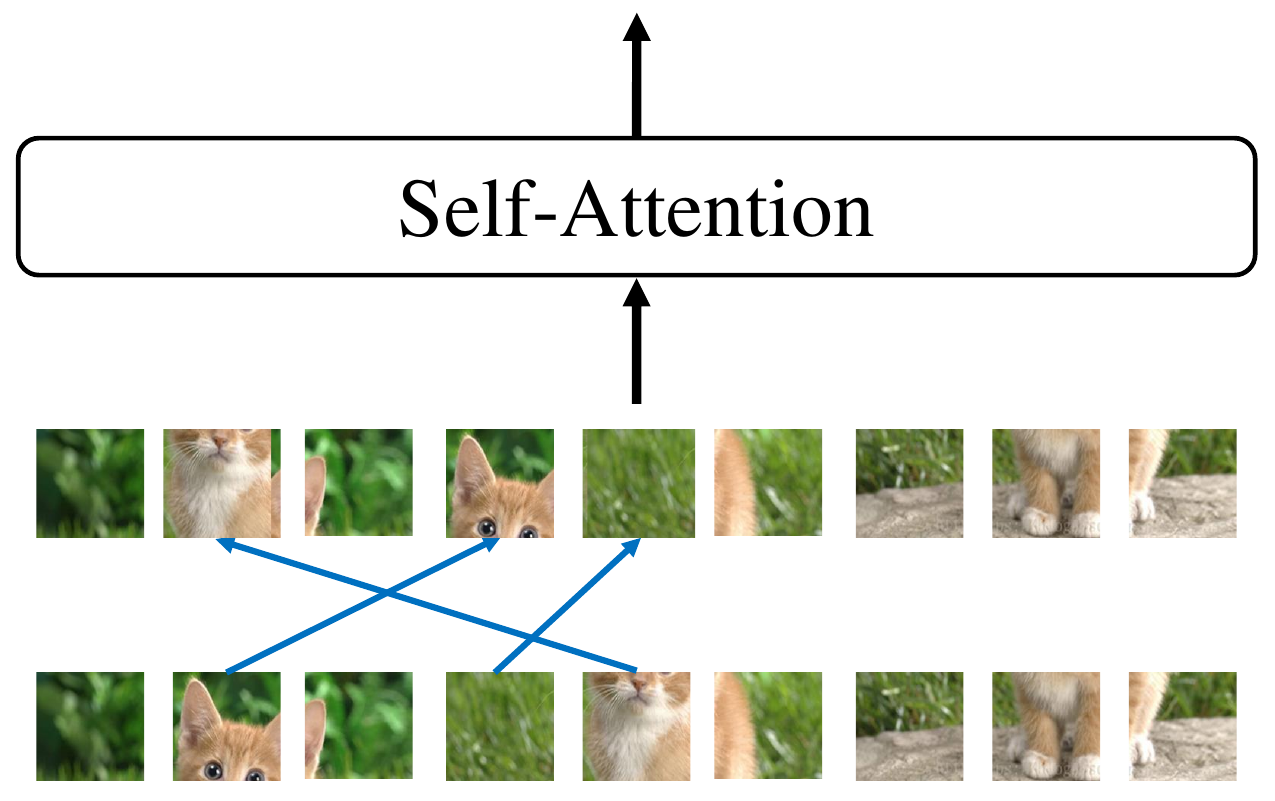}
    \caption{\footnotesize Random Patch Permutation.}
    \label{fig:vit_rand1}
    \end{subfigure} \hspace{0.05cm}  
    \begin{subfigure}[b]{0.3\textwidth}
    \centering
    \includegraphics[scale=0.15]{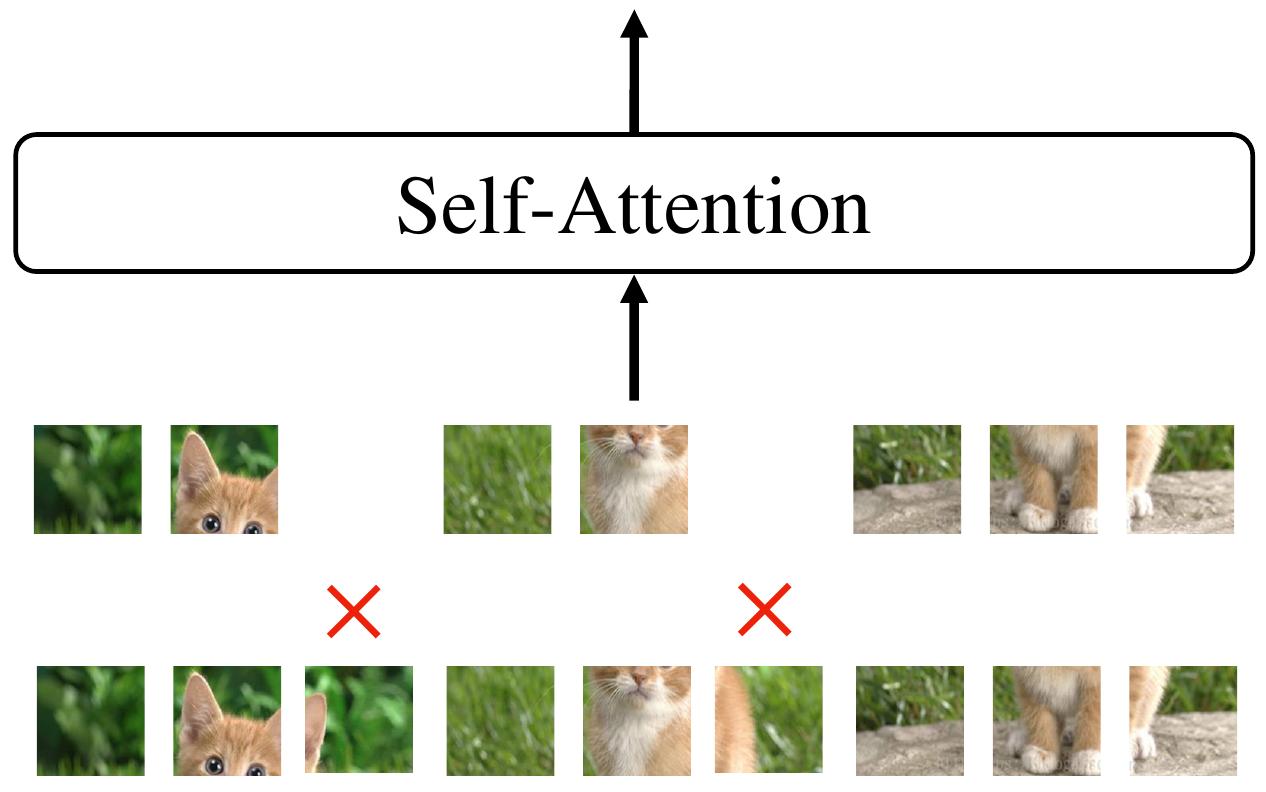}
    \caption{\footnotesize Random Patch Drop.}
    \label{fig:vit_rand2}
    \end{subfigure} \hspace{0.08cm}  
    \begin{subfigure}[b]{0.36\textwidth}
    \centering
    \includegraphics[scale=0.2]{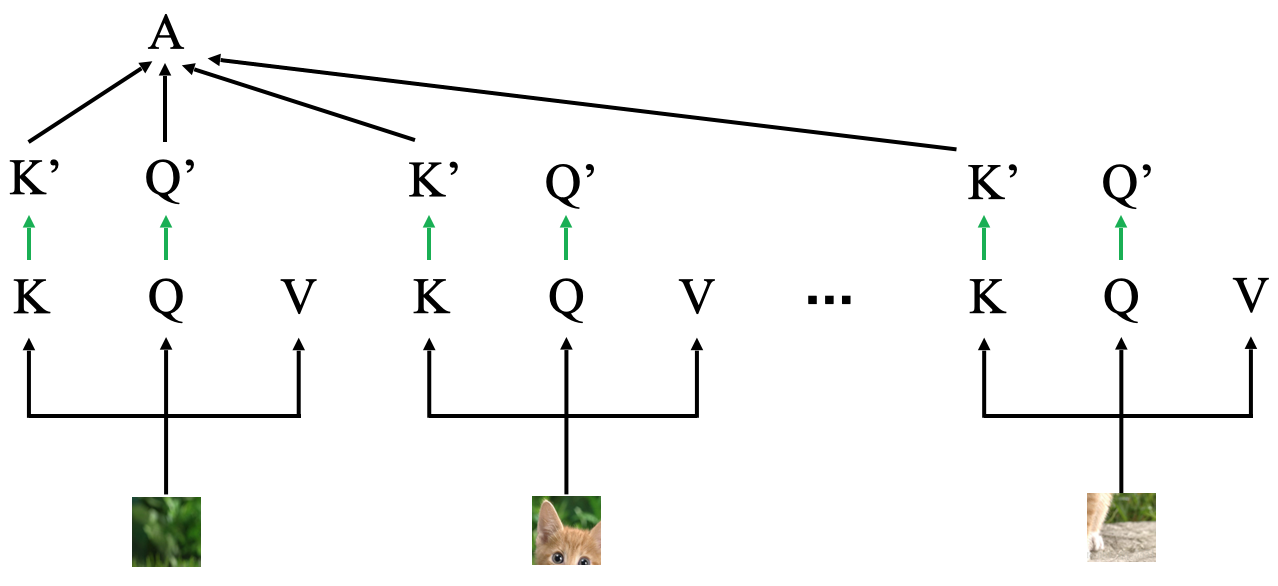}\vspace{-0.1cm} 
    \caption{\footnotesize Randomness in Attention.}
    \label{fig:vit_rand3}
    \end{subfigure}

    \caption{Non-additive Randomness in ViT. The color marks denote our operations to introduce randomness into ViT. The blue arrows in (a) are the random permutation operation. The red crosses in (b) are the random drop operation. The green arrows in (c) are the random dimension reduction operation. All three operations introduce non-additive randomness.}
    \label{fig:vit_rand}
\end{figure}

\subsubsection{Patch Sequence Random Permutation in ViT}
The self-attention module in ViT models sequence to sequence. In the standard ViT, the input sequence of image patches is 
$
\boldsymbol{x} = \{\boldsymbol{x}_i \vert i \in (1,2,3, ..., {HW/P^2}) \}.
$
To encode positional information of input patches, ViT add explicit positional embedding to the corresponding patch embedding. The positional embedding can be 
$
\boldsymbol{p} = \{\boldsymbol{p}_i \vert i \in (1,2,3, ..., {HW/P^2}) \}.
$
The initial patch embedding of the $i$-th patch is composed of the embedding of a raw patch $\boldsymbol{x}_i$ and the corresponding positional embedding of $\boldsymbol{p}_i$, specifically, the sum of the embeddings $E(\boldsymbol{x}_i) + E(\boldsymbol{p}_i)$. The sum operation encodes the positional information into patch embedding.

We introduce randomness into the position information encoding process. We first randomly sample parts of patches and permute their correspondence to their positional embedding. In other words, once selected, the initial patch embedding of the $i$-th patch is $E(\boldsymbol{x}_i) + E(\boldsymbol{p}_j)$ where $\boldsymbol{p}_j$ is the positional embedding of the $j$-th patch. Both the patch selection and the correspondence permutation are different from run to run. 

The self-attention operation itself is invariant to sequence order. Hence, our patch permutation is equivalent to the permutation of the input patches of ViT model, as shown in Fig.~\ref{fig:vit_rand1}. The probability of selecting patches to permute controls the trade-off between the defensive effect and the clean performance.

\subsubsection{Patch Random Drop in ViT}
It has been said that ViT has less inductive bias than traditional CNNs~\cite{dosovitskiy2020image}. This is because only grid-structured input is acceptable for standard CNNs in image classification tasks. Contrarily, given its architectural traits, the self-attention module of ViT can accept sequences with different lengths. Given the architectural flexibility, we introduce non-additive randomness by randomly sampling the elements of the input sequence. Namely, the input sequence of image patches is
$
\boldsymbol{x}' = \{ \phi_r(\boldsymbol{x}_i, p) \vert i \in (1,2,3, ..., {HW/P^2}) \},
$ where $p$ is a pre-defined probability and the function $f_r(\boldsymbol{x}_i, p)$ is defined as
$$ \phi_r(\boldsymbol{x}_i, p) = 
\begin{cases}
\boldsymbol{x}_i\hspace{0.5cm} \text{if } u > p\\
\{\} \hspace{0.5cm} \text{if } u \leq p,
\end{cases}
$$ where u is a value sampled from the uniform distribution $\mathcal{U}(0, 1)$.
In the non-additive randomness above, different image patches are sampled in every forward pass, as shown in Fig.~\ref{fig:vit_rand2}. Namely, only part of the inputs are taken to make a prediction, meanwhile, the selected parts are random in each prediction. Given its remarkable modeling capacity, ViT still performs well when such non-additive randomness is introduced.

\subsubsection{Patch Attention Perturbation by Sampling Keys and Queries}
Another non-additive randomness we explore is in the process to create attention. The attention of the $i$-th patch is computed with the $i$-th  query $\boldsymbol{Q}^{l+1}_i$ and the keys $\boldsymbol{K}^{l+1}$ of all patches. We introduce randomness into attention by injecting it into the query and the keys. 

The key and the query of a patch are represented by activation vectors of the same dimension. Given the dimension $D$ of the query and the keys, their dimension indices are $I = (1, 2, ..., D)$. As illustrated in Fig.~\ref{fig:vit_rand3}, we first sample the dimension indices to obtain I' where $\vert I \vert  \leq \vert I' \vert$. We drop each dimension with a probability of $\alpha \in [0, 1)$. We propose to formulate the new keys and the queries with the sampled dimensions $I'$.
$
\boldsymbol{K}'_i = \psi(\boldsymbol{K}_i, \; I'),  \;\;\;\; \boldsymbol{Q}'_i = \psi(\boldsymbol{Q}_i, \; I'),
$ where $\psi(\cdot)$ reduces dimension by dropping the specified dimensions. Namely, only the sampled dimensions are considered. By doing this, we introduce non-additive randomness into attention. The new attention is computed with the new keys and the queries.

The non-additive model randomness is different from the input with additive robustness where the input sequence is 
$
\boldsymbol{x} = \{\boldsymbol{x}_i +  \mathcal{N}(\boldsymbol{0},  \sigma^2\boldsymbol{I}
) \vert i \in (1,2,3, ..., {HW/P^2}) \},
$
where $\sigma$ is the variance of gaussian noise.

\subsection{Analysis of (Non)-additive Randomness for Adversarial Defense}
In this subsection, we first show why stochastic defense strategies are effective against QBBA. Furthermore, we provide an analysis to show why non-additive randomness on the model performs better than other stochastic defense strategies.

\subsubsection{Defense against Optimization-based Black-box Attacks}
Optimization-based black-box attacks~\cite{chen2017zoo,cheng2018query,cheng2019sign} create adversarial examples using estimated gradients. The gradient estimation is often implemented with finite difference~\cite{chen2017zoo}. Given the model $f(\boldsymbol{x}, \boldsymbol{\theta})$ with the input $\boldsymbol{x} \in \mathbb{R}^{(H \times W \times C )}$ and the model parameters $\boldsymbol{\theta}$, the input gradients $\boldsymbol{G} \in \mathbb{R}^{(H \times W \times C )}$ can be estimated as follows:
\vspace{-0.1cm}
\begin{equation}
\begin{gathered}
    \boldsymbol{G} = \frac{1}{N}\sum_{i=0}^N \frac{f(\boldsymbol{x} +\beta \boldsymbol{u}_i, \; \boldsymbol{\theta}) - f(\boldsymbol{x} -\beta \boldsymbol{u}_i, \; \boldsymbol{\theta})}{2\beta}
\end{gathered}
\label{eqn:gradient}
\end{equation}
where $\boldsymbol{u}_i \sim \mathcal{N}(\boldsymbol{0}, \sigma^2\boldsymbol{I})$ and $\boldsymbol{G}=\frac{1}{N}\sum_{i=0}^N \boldsymbol{G}_i$.
The stochastic defense strategies introduce randomness to inputs or models. The gradient estimation based on the perturbed inference can differ from the original one. Note that the randomness $\boldsymbol{\eta}$ and $\boldsymbol{\eta}'$ in the forward inference are different from run to run. The introduced randomness leads to inaccurate gradients:
\begin{equation}
 \boldsymbol{\hat{G}}_i=\frac{f(\boldsymbol{x} +\beta \boldsymbol{u}_i, \boldsymbol{\theta}, \; \boldsymbol{\eta}) - f(\boldsymbol{x}-\beta \boldsymbol{u}_i, \boldsymbol{\theta}, \; \boldsymbol{\eta}')}{2\beta} \neq \boldsymbol{G}_i.
\end{equation}

Concretely, when the additive randomness is added to input, the forward inference is $f(\boldsymbol{x} +\beta \boldsymbol{u}+ \boldsymbol{\eta}, \; \boldsymbol{\theta})$. In case $\boldsymbol{\eta}=-2\beta \boldsymbol{u}$, $\boldsymbol{\eta}'=2\beta \boldsymbol{u}$, the gradients can even be $sign(\boldsymbol{\hat{G}}_i) = - sign(\boldsymbol{G}_i)$, which leads to total wrong estimations. In practice, the estimated gradients are often opposite to the original ones in some dimensions of $\boldsymbol{G}_i$. When the additive randomness is added to model parameters, the output of the perturbed layer is $(\boldsymbol{w}+\boldsymbol{\eta})\boldsymbol{x} + \boldsymbol{b} =\boldsymbol{wx}+\boldsymbol{\eta x} + \boldsymbol{b}$, which is equivalent to adding perturbations to the input of the next layer.

When the non-additive randomness is added to input (e.g. affine transformation), the forward inference is $f(g(\boldsymbol{x}), \; \boldsymbol{\theta})$. The finite difference does work well in this case since the two inputs in two forward inferences differ from each other significantly. Our proposed method is to add non-additive randomness to the model, i.e., sampling model functions in forwarding inferences. The forward inference $f'(\boldsymbol{x}, \; \boldsymbol{\theta})$ makes the gradient estimation inaccurate by changing the model function. The architectural flexibility of ViT models makes it feasible to introduce non-additive randomness to the model.

\subsubsection{Defense Against Search-based Black-box Attacks}
Search-based black-box attacks first add random noises $\boldsymbol{u} = \{\boldsymbol{u}_k \vert k \in (1,2, ..., N) \}$ to the current input, then select the noise that leads to the maximal loss $L_k = \ell(f(\boldsymbol{x} + \boldsymbol{u}_k, \; \boldsymbol{\theta}))$, and repeat the first two steps until an effective adversarial perturbation is found. When the additive robustness $\boldsymbol{\eta} = \{\boldsymbol{\eta}_i \vert i \in (1,2, ..., M) \}$ is added to the input, the loss is $L'_k = \ell(f(\boldsymbol{x} + \boldsymbol{u}_k + \boldsymbol{\eta}_i, \; \boldsymbol{\theta}))$ where $i$ is randomly selected from $\{1,2, ..., M\}$ in each forward pass. The new loss makes search-based methods infeasible since $L'_k$ is not fixed from run to run due to $\boldsymbol{\eta}_i$. Note that $\boldsymbol{\theta}_i$ is introduced by the defense side, which is unavailable to the adversary.

Similarly, when the non-additive randomness is introduced to input, the loss becomes $L'_k = \ell(f(g(\boldsymbol{x} + \boldsymbol{u}), \; \boldsymbol{\theta}))$. The selected adversarial perturbation $\boldsymbol{u}_i$ does not necessarily lead to a higher loss in the next search iteration. Furthermore, the non-additive randomness on the model changes the model function. The $\boldsymbol{x} + \boldsymbol{u}$ is able to fool the model function $f'$, but not necessarily another one $f''$. This error in the adversarial perturbation selection makes the searching process almost similar to a random walk.

\subsubsection{Adaptive Attacks on Stochastic Defense}
In the setting above, the adversary has no access to the model details as well as the defense strategies for a black-box attack setting. We now consider the black-box attack attacks that also work on the model with stochastic defense, i.e. defense-aware attacks. As suggested in~\cite{athalye2018obfuscated}, a defense-aware attack is to combine the Expectation of Transform strategy (EOT) with the existing black-box attacks. We argue that defense-aware attacks do not work well for the following two reasons: The estimated gradients are still wrong under EOT attack strategy since the models are highly non-linear~\cite{byun2022effectiveness}; The application of EOT requires a large number of queries on the model, which is what the adversary tries to avoid in the first place when QBBAs are designed.

\subsubsection{Efficiency of Defense Strategy}
Besides the effectiveness of the defense strategies, their efficiency is also important since some defense strategies can bring large extra computational costs. Concretely, adversarial training is very computationally expensive in the training stage~\cite{madry2017towards}. The ensemble-based defense strategies increase the cost many times more~\cite{liu2018towards}. The recently proposed randomness-based defense strategies only bring tiny extra computational costs. Different from the previous ones, our approach can even accelerate the forward inference of ViT, such as Patch Random Drop and Patch Attention Perturbations. Our non-additive randomness on the model also enjoys the advantage of high efficiency.

\section{Experiments}
We first briefly describe the baselines and our approach. Then, we show the experimental setting as well as evaluation metrics. Next, we show our experimental results and analysis.

\textbf{Methods.} The following defense methods have been applied in our experiments: Small Noise Defense (\textbf{SND})~\cite{byun2022effectiveness}, Parametric Noise Injection (\textbf{PNI})~\cite{he2019parametric}, Random Resizing and Padding (\textbf{R\&P})~\cite{xie2017mitigating}, Patch Random Permutation (\textbf{PRPerm}), Patch Random Drop (\textbf{PRDrop}), and Patch Attention Perturbation (\textbf{PAttnPert}). We select representative QBBAs, such as Square Attack~\cite{andriushchenko2020square}, Hop Skip Jump Attack(HSJA)~\cite{chen2020hopskipjumpattack}, and GeoDA~\cite{rahmati2020geoda}. The details of all attack and defense methods can be found in Supplement B and C.

\textbf{Experimental Setting.} We use the most primary ViT models~\cite{dosovitskiy2020image,touvron2021training}. The comparison to various CNN models (e.g. ResNet) and the generalization on other ViT models (e.g. SwinTransformer) are also studied in our experiments. All models we used are from Timm library~\cite{rw2019timm}, which are pre-trained on the ImageNet dataset~\cite{rw2019timm}. Following most previous work on black-box attacks, we take 1k images from the validation dataset as the test set.

We select the popular QBBA to compare different defense approaches. The hyper-parameters of each attack are shown in Supplement B. Instead of carefully selecting a hyper-parameter, we report the robust accuracy and clean accuracy under different hyper-parameters of defense strategies. By doing this, we compare the trade-off achieved by different approaches.

\textbf{Evaluation Metrics.} As discussed before, there is a trade-off between robust accuracy and clean accuracy. For a fair comparison, we report the clean accuracy (\textbf{Accu} in \%) on the whole validation dataset as well as Attack Failure Rate (\textbf{AFR} in \%) on the selected image set. When the attack is unable to fool the target model under the allowed perturbation range within the given number of attack iterations, the attack is counted as a failure.

\begin{figure}[!t]
\centering
    \begin{subfigure}[b]{0.32\textwidth}
    \centering
    \hspace{-0.3cm}\includegraphics[scale=0.4]{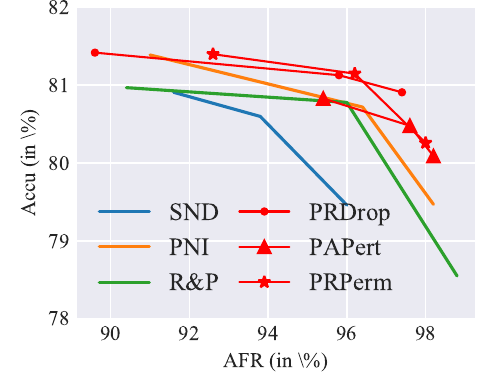}\vspace{-0.1cm}  
    \caption{\footnotesize Square Attack~\cite{andriushchenko2020square}}
    \label{fig:vit_square}
    \end{subfigure} 
    \begin{subfigure}[b]{0.32\textwidth}
    \centering
    \hspace{-0.16cm}\includegraphics[scale=0.4]{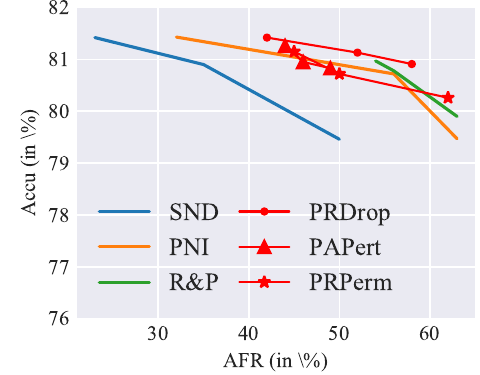}\vspace{-0.1cm} 
    \caption{\footnotesize HSJA~\cite{chen2020hopskipjumpattack}}
    \label{fig:vit_hsja}
    \end{subfigure}
    \begin{subfigure}[b]{0.32\textwidth}
    \centering
    \hspace{-0.14cm}\includegraphics[scale=0.4]{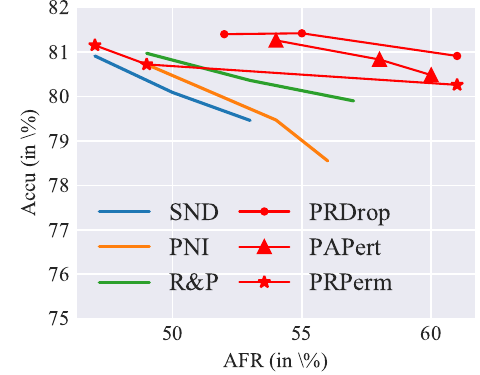}
    \caption{\footnotesize GeoDA~\cite{rahmati2020geoda}}
    \label{fig:vit_geoda}
    \end{subfigure}

    \caption{Stochastic Defense on ViT against Query-based Black-box Attacks. In each subfigure, the x-axis and y-axis are the clean accuracy (Accu in \%) and the attack failure rate (AFR in \%), respectively. Each of the lines corresponds to a type of defense. The three red lines are our non-additive randomness defense on the model. Each point in the line corresponds to a trade-off point between Accu and AFR. Our methods can achieve a better trade-off than others.}\vspace{-0cm}
    \label{fig:vit_defense}
\end{figure}

\subsection{Stochastic Defense}
We equip the ViT-small models with different defense methods respectively and apply the state-of-the-art QBBA to the models. We report the results in Fig.~\ref{fig:vit_defense}. Each subfigure shows the results of one type of attack. For example, the subfigure~\ref{fig:vit_square} presents the results on Square Attack. The x-axis and y-axis are the clean accuracy (Accu in \%) and the attack failure rate (AFR in \%), respectively. In each subfigure, there are six lines, each of which corresponds to a type of defense method. Concretely, the blue, green, and yellow lines correspond to SND, R\&P, and PNI defense methods, respectively. The three red lines are our method, where each mark corresponds to a method under our defense based on the non-additive randomness of the model. Each point in the line corresponds to a trade-off point between Accu and AFR.

In the subfigure~\ref{fig:vit_square}, we can observe that our method can achieve better defense (i.e. the higher AFR) when keeping the same clean accuracy. Overall, our methods marked with red lines can achieve a better trade-off than the others. The claim is also true when different QBBA are applied. The experimental results on more QBBAs are in Supplement A, such as NES Attack~\cite{ilyas2018black}, SimBA~\cite{guo2019simple}, and Boundary~\cite{brendel2017decision}.

\begin{figure}[!t]
\centering
    \begin{subfigure}[b]{0.32\textwidth}
    \centering
    \hspace{-0.3cm}\includegraphics[scale=0.4]{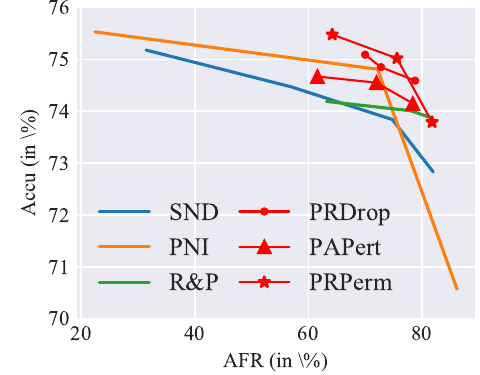}
    \caption{\footnotesize ViT Tiny}
    \end{subfigure} 
    \begin{subfigure}[b]{0.32\textwidth}
    \centering
    \hspace{-0.16cm}\includegraphics[scale=0.4]{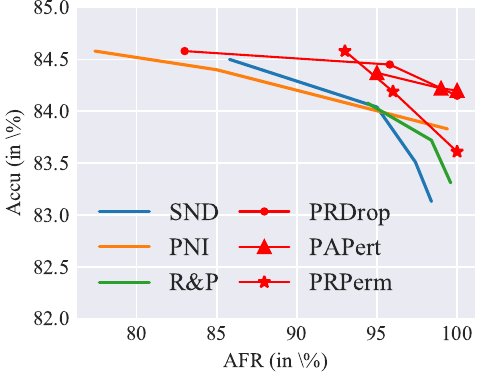}
    \caption{\footnotesize ViT Base}
    \end{subfigure}
    \begin{subfigure}[b]{0.32\textwidth}
    \centering
    \hspace{-0.14cm}\includegraphics[scale=0.4]{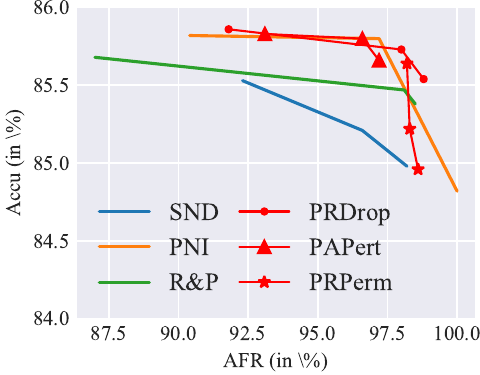}
    \caption{\footnotesize ViT Large}
    \end{subfigure} \vspace{-0.2cm}
    \caption{Stochastic Defense on ViTs with different model sizes. Besides ViT-small, we also conduct experiments on ViT-tiny, ViT-base, and ViT-large. Our non-additive stochastic defense on the model (represented by the three red lines) can achieve a better trade-off than other defense methods.}
    \label{fig:vit_defense_sizes}
\end{figure}

\textbf{Stochastic Defense on Different Models.}
We also investigate the impact of model sizes and model architectures on the defense methods. Besides the ViT-small, we test the defense methods on ViT-tiny, ViT-base, and ViT-huge with the state-of-the-art Square Attack. The results are in Fig.~\ref{fig:vit_defense_sizes}. Our defense methods outperform others on ViT models with different sizes.

\begin{figure}[!t]
\centering
    \begin{subfigure}[b]{0.32\textwidth}
    \centering
    \hspace{-0.3cm}\includegraphics[scale=0.4]{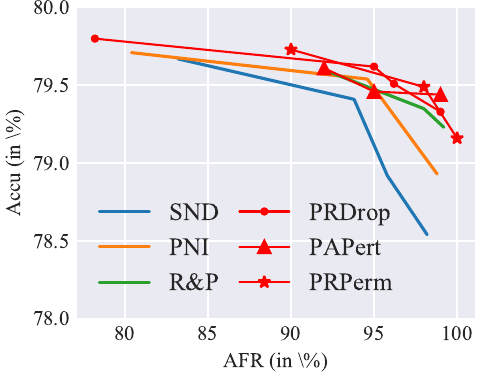} 
    \caption{\footnotesize DeiT~\cite{touvron2021training}}
    \end{subfigure}\hspace{-0.1cm}
    \begin{subfigure}[b]{0.32\textwidth}
    \centering
    \includegraphics[scale=0.4]{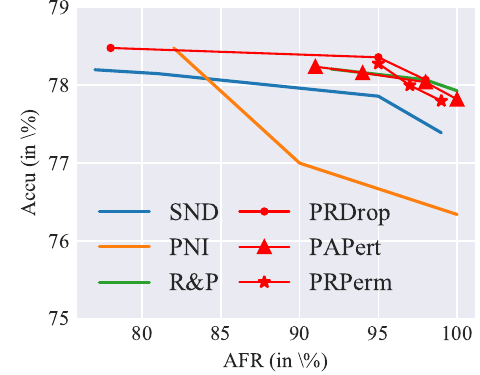} 
    \caption{\footnotesize CaiT~\cite{rajasegaran2019deepcaps}}
    \end{subfigure}
    \begin{subfigure}[b]{0.32\textwidth}
    \centering
    \hspace{-0.05cm}\includegraphics[scale=0.4]{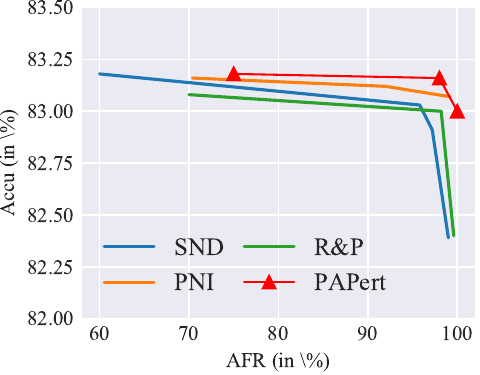}
    \caption{\footnotesize Swin Transformer~\cite{liu2021swin}}
    \end{subfigure} \vspace{-0.2cm}
    \caption{We apply the stochastic defense to models with different model architectures. On the popular versions of Vision Transformers, our methods (marked with red lines) are superior to others.}
    \label{fig:vit_defense_archi}
\end{figure}

There are several popular ViT variants. In this experiment, we select some representative architectures, namely, DeiT, CaiT, and SwinTransformer. We apply various defenses on the selected models and test the defended models with Square Attack. Note that the patch random drop approach cannot be directly applied to Swin Transformer since the shifting and merging operating therein make the patch drop challenging. The results are reported in Fig.~\ref{fig:vit_defense_archi}. Our methods perform better than others on different models with different architectures.

\textbf{Combination of Defense Strategies.}
We also investigate the combinations of different stochastic defense methods. As shown in Supplement D, the combination is still effective when we combine three types of the proposed methods. In other words, the defense is similarly effective when we randomly drop and permute the patches and perturb the attention at the same time.

\subsection{Trade-off between Robust Accuracy and Clean Accuracy}
The main limitation of the randomization defense method is the decreased model performance on clean examples. If the injected randomness is too small, the defensive effect is low. In contrast, the large noise injection leads to better defensive effectiveness, but low model performance on clean examples. Our method achieves a good trade-off, where the non-additive randomness-based defense strategy hardly reduces the clean performance of ViT. For example, the ViT-small has a clean accuracy of 81.4\%. When equipped with our defense strategy, it shows a robust accuracy of 92.6\% and also the same clean accuracy of 81.4\%.

We now discuss the possibility of further improving the trade-off. The intuitive solution is the finetuning method. A pre-trained model is finetuned to improve clean performance where the model is modified with the defense strategy. In other words, we use the same model during finetuning as in the inference. The finetuning strategy can be applied to boost the model for both additive and non-additive randomization. The finetuned models with improved clean accuracy can achieve a better trade-off than the un-fintuned counterparts. For example, the ViT-small with our non-additive randomness can achieve the robust accuracy of 98.4\% and the clean accuracy of 81.4\% after finetuning.

In the defense of our non-additive randomness, the model performance will be reduced due to the reduced number of input patches or less accurate attention. Besides the finetuning method, another way to alleviate the model degradation is with scheduled non-additive randomness. Note that it is also possible to apply the proposed non-additive randomness to a specific layer instead of all layers. We leave more explorations of more sophisticated fine-tuning methods in future work.

\begin{table}[t]
\begin{center}
\footnotesize	
\setlength\tabcolsep{0.25cm}
\begin{tabular}{c | c | c | c | c | c | c | c}
\toprule
Models   & Defense   & Accu(\%) & AFR(\%) & Models    & Defense   & Accu(\%) & AFR(\%) \\
\midrule
\multirow{4}{*}{ResNet18} & No  & 69.55  & 0.0  & \multirow{4}{*}{ResNet50} & No  & 75.86  & 0.7  \\
 & SND~\cite{byun2022effectiveness}  & 69.21  & 43.1  &  & SND~\cite{byun2022effectiveness}  & 75.79  & 36.2  \\
 & PNI~\cite{he2019parametric}  & 69.45  & 38.0  &  & PNI~\cite{he2019parametric}  & 75.84  & 32.3  \\
 & R\&P~\cite{xie2017mitigating}  & 69.35  & 68.3  &  & R\&P~\cite{xie2017mitigating}  & 75.22  & 78.5  \\
\midrule
\multirow{7}{*}{ViT-tiny} & No  & 75.48  & 0.9  & \multirow{7}{*}{ViT-small} & No  & 81.40  & 1.2  \\
 & SND~\cite{byun2022effectiveness}  & 75.18  & 41.14  &  & SND~\cite{byun2022effectiveness}  & 81.38  & 72.0  \\
 & PNI~\cite{he2019parametric}  & 74.81  & 72.2  &  & PNI~\cite{he2019parametric}  & 81.40  & 33.5  \\
 & R\&P~\cite{xie2017mitigating}  & 74.19  & 63.2  &  & R\&P~\cite{xie2017mitigating}  & 80.97  & 89.5  \\
 \cmidrule{2-4} \cmidrule{6-8}
 & PRDrop  & 75.09  & 70.0  &  & PRDrop  & 81.39  & 90.6  \\
 & PAPert  & 74.55  & 72.1  &  & PAPert  & 80.98  & 95.4  \\
 & PRPerm & 75.48  & 64.2  &   & PRPerm  & 81.40  & 92.6  \\
\bottomrule  
\end{tabular} \vspace{0.3cm}
\caption{Comparison of ViT with ResNet for Defending Against Squared Attack. Vit with the randomness-based defense achieves a better trade-off than the counterpart ResNet (e.g., ResNet50 vs. ViT-small). Besides, our non-additive randomness-based defense on ViT achieves a better trade-off than the other randomness-based defense methods.} \vspace{-0.5cm}
\label{tab:comp_cnn}
\end{center}
\end{table}

\subsection{Comparison with ResNet for Defense}
We now study the question of whether ViT achieves a better trade-off than ResNet with or without our defense. The results are reported in Tab.~\ref{tab:comp_cnn}. ViT achieves a better trade-off than ResNet to defend against the state-of-the-art QBBA.

Given the flexibility of ViT architectures, we can drop and permutate the input and apply non-additive randomness to its attention. However, it is less easy to do so in CNNs since CNN requires input with strict grid data structures. One of our non-additive randomness-based strategies is to permute the input patches of each self-attention module. A similar defense strategy can also be applied to ResNet where we randomly select input patches and permute them. Our experiments show patch permutation does not work well on ResNet. We conjecture that the defense pattern of a patch is local, which is still effective when placed in different places.

\vspace{-0.3cm}
\section{Conclusion}
Random-based defense strategies have been intensively explored to defend against various black-box attack methods. In this work, we taxonomize the defensive randomness from the perspective of defense against query-based black-box attacks. Following our taxonomy, we propose non-additive randomness on ViT. Specifically, we propose three ways to integrate non-additive robustness in the self-attention module of ViT. Our experiments verify that our defense method can achieve better trade-offs than the rest of the stochastic methods in our taxonomy. Based on the non-additive randomness, this work boosts the robustness of ViT to query-based black-box attacks. Almost all recent foundation models are transformer architecture-based. Extending our proposed techniques from classifiers to those foundation models remains to explore for improving their robustness~\cite{chen2023benchmarking,gu2023towards}. We leave further exploration in future work.

\noindent\textbf{Acknowledgements}
This work is partially supported by the UKRI grant: Turing AI Fellowship EP/W002981/1 and EPSRC/MURI grant: EP/N019474/1. We would also like to thank the Royal Academy of Engineering and FiveAI.

\bibliography{egbib}
\newpage
\appendix
\section*{A: Experiments on More Query-based Black-box Attacks}
\vspace{-0.3cm}
\begin{figure}[!h]
\centering
    \begin{subfigure}[b]{0.32\textwidth}
    \centering
    \hspace{-0.3cm}\includegraphics[scale=0.5]{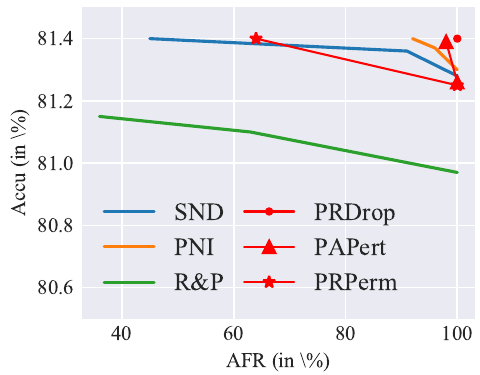}\vspace{-0.1cm}  
    \caption{\footnotesize NES Attack}
    \label{fig:vit_square}
    \end{subfigure} 
    \begin{subfigure}[b]{0.32\textwidth}
    \centering
    \hspace{-0.16cm}\includegraphics[scale=0.5]{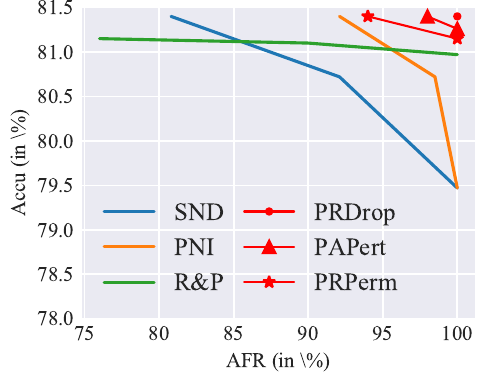}\vspace{-0.1cm} 
    \caption{\footnotesize SimBA}
    \label{fig:vit_hsja}
    \end{subfigure}
    \begin{subfigure}[b]{0.32\textwidth}
    \centering
    \hspace{-0.14cm}\includegraphics[scale=0.5]{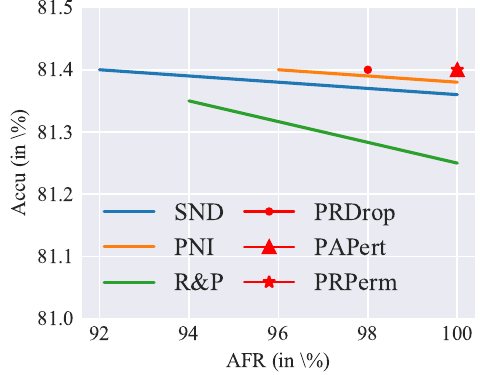}\vspace{-0.1cm} 
    \caption{\footnotesize Boundary}
    \label{fig:vit_geoda}
    \end{subfigure}

    \caption{Randomness-based Defense on ViT-small against Query-based Black-box Attacks. Besides SOTA attacks, we provide the results on more attack methods. In each subfigure, the x-axis and y-axis are the clean accuracy (Accu in \%) and the attack failure rate (AFR in \%), respectively. Each of the lines corresponds to a type of defense. The three red lines are our non-additive randomness defense on the model. Each point in the line corresponds to a trade-off point between Accu and AFR. Our methods can achieve a better trade-off than others.}
\end{figure}\vspace{-0.1cm}
\begin{figure}[!h]
\centering
    \begin{subfigure}[b]{0.32\textwidth}
    \centering
    \hspace{-0.3cm}\includegraphics[scale=0.5]{figures/vit_tiny_Squared.pdf}\vspace{-0.1cm}  
    \caption{\footnotesize Square Attack}
    \label{fig:vit_square}
    \end{subfigure} 
    \begin{subfigure}[b]{0.32\textwidth}
    \centering
    \hspace{-0.16cm}\includegraphics[scale=0.5]{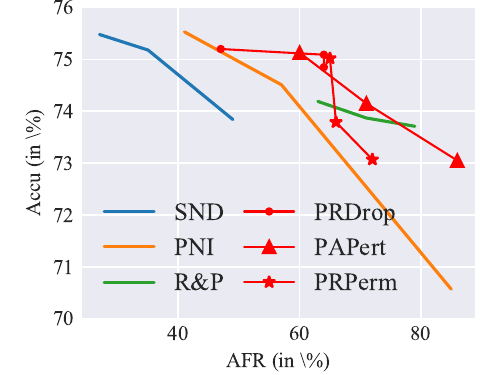}\vspace{-0.1cm} 
    \caption{\footnotesize HSJA}
    \label{fig:vit_hsja}
    \end{subfigure}
    \begin{subfigure}[b]{0.32\textwidth}
    \centering
    \hspace{-0.14cm}\includegraphics[scale=0.5]{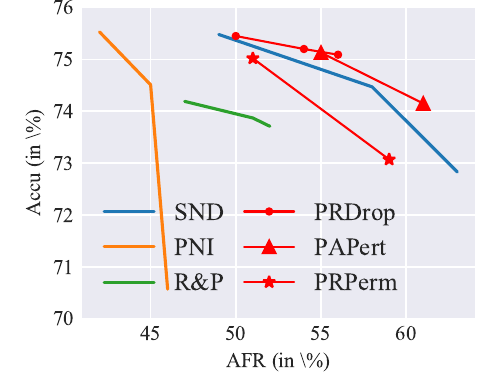}\vspace{-0.1cm} 
    \caption{\footnotesize GeoDA}
    \label{fig:vit_geoda}
    \end{subfigure}

    \caption{Randomness-based Defense on ViT-tiny against Query-based Black-box Attacks. We also provide the results on ViT models with different sizes. Our methods still achieve a better trade-off than others.}
\end{figure}

\section*{B: Hyper-parameters of Query-based Black-box Attacks}

\begin{table}[!h]
\caption{Hyper-parameters of Square Attack \cite{andriushchenko2020square}}
\label{tab:fair_models}
\begin{center}
\footnotesize
\setlength\tabcolsep{0.6cm}
\begin{tabular}{ccc}
\toprule
Maximum number of iterations   & 100 \\
\midrule
Maximum perturbation   & 0.03 \\
\midrule
Initial fraction of elements   & 0.8 \\
\midrule
Number of restarts   & 1 \\
\bottomrule
\end{tabular}
\end{center}
\end{table}

\begin{table}[!h]
\caption{Hyper-parameters of HSJA \cite{chen2020hopskipjumpattack}}
\label{tab:fair_models}
\begin{center}
\footnotesize
\setlength\tabcolsep{0.6cm}
\begin{tabular}{cc}
\toprule
Maximum number of iterations   & 20 \\
\midrule
Maximum number of evaluations for estimating gradient   & 10000 \\
\midrule
Initial number of evaluations for estimating gradient  & 100 \\
\midrule
Maximum number of trials for initial generation of AE   & 100 \\
\bottomrule
\end{tabular}
\end{center}
\end{table}

\begin{table}[!h]
\caption{Hyper-parameters of GeoDA \cite{rahmati2020geoda}}
\label{tab:fair_models}
\begin{center}
\footnotesize
\setlength\tabcolsep{0.6cm}
\begin{tabular}{ccc}
\toprule
Dimensionality of 2D frequency space (DCT)   & 20 \\
\midrule
Maximum number of iterations   & 4000 \\
\midrule
binary search tolerance   & 0.0001 \\
\midrule
Variance of the Gaussian perturbation  & 0.0002 \\
\bottomrule
\end{tabular}
\end{center}
\end{table}

\begin{table}[!h]
\caption{Hyper-parameters of NES Attack \cite{ilyas2018black}}
\label{tab:fair_models}
\begin{center}
\footnotesize
\setlength\tabcolsep{0.6cm}
\begin{tabular}{ccc}
\toprule
Maximum number of trials per iteration   & 1000 \\
\midrule
Maximum perturbation   & 0.05 \\
\midrule
Maximum number of evaluations for estimating gradient   & 100 \\
\midrule
Variance of the Gaussian perturbation  & 0.001 \\
\bottomrule
\end{tabular}
\end{center}
\end{table}

\begin{table}[!h]
\caption{Hyper-parameters of SimBA \cite{guo2019simple}}
\label{tab:fair_models}
\begin{center}
\footnotesize
\setlength\tabcolsep{0.6cm}
\begin{tabular}{ccc}
\toprule
Norm of Perturbation   & L2 norm \\
\midrule
Maximum Perturbation   & 0.2 \\
\midrule
Dimensionality of 2D frequency space (DCT)   & 40 \\
\midrule
Maximum number of iterations   & 10000 \\
\midrule
ordering for coordinates & random \\
\bottomrule
\end{tabular}
\end{center}
\end{table}

\begin{table}[!h]
\caption{Hyper-parameters of Boundary \cite{brendel2017decision}}
\label{tab:fair_models}
\begin{center}
\footnotesize
\setlength\tabcolsep{0.6cm}
\begin{tabular}{ccc}
\toprule
Initial step size for the orthogonal step   & 0.01 \\
\midrule
Initial step size for the step towards the target   & 0.01 \\
\midrule
Factor by which the step sizes are multiplied   & 0.667 \\
\midrule
Maximum number of iterations   & 5000 \\
\midrule
Maximum number of trials per iteration   & 25 \\
\midrule
Number of samples per trial   & 20 \\
\midrule
Maximum number of trials for initial generation of AE  & 100 \\
\midrule
Stop attack if perturbation is smaller than  & 0 \\
\bottomrule
\end{tabular}
\end{center}
\end{table}

\clearpage
\section*{C: Details and Setting of Defense Methods}
The details and setting of the baselines and our approaches are shown as follows.

\textbf{Small Noise Defense (SND).}~\cite{byun2022effectiveness} proposes to defend against query-based black-box attacks by adding random Gaussian noise to inputs. The variance of the gaussian noise controls the strength of the noise, which balances the trade-off between robust accuracy and clean accuracy. We report the multiple results on different variance values, which are in $[0.001, 0.1]$.

\textbf{Parametric Noise Injection (PNI).}~\cite{he2019parametric} proposes to add layer-wise trainable Gaussian noise to the activation or weight of each layer. In this work, we add random Gaussian noise to activations of all layers without retraining to keep a fair comparison. Only the inference stage is changed. Similarly, the variance of the Gaussian noise balances the trade-off between robustness and clean performance. The variance values we use are in $[0.001, 0.2]$.

\textbf{Random Resizing and Padding (R\&P).}~\cite{xie2017mitigating} proposes a pre-processing-based defense method where the input is randomly resized and padded. The random resize and padding are different in different forward passes. The input image with a fixed size (i.e. $224\times224$) is resized to a smaller size (e.g. $208\times208$) and padded to the original size. The downsampled sizes control the trade-off between robust accuracy and clean accuracy, which are in $[160, 218]$.

\textbf{Patch Random Permutation (PRPerm).} In our PRP, the positional embedding of a certain percentage of patches is randomly permuted. The permutation of positional embedding is equivalent to the permutation of the corresponding input patches. The percentage of the permuted patches controls the trade-off. That too many patches are permuted leads to better robustness, but unsatisfying performance on clean inputs. Note that the permutation only happens before the first self-attention module where positional embedding is available. The percentage values we use are in $[1\%, 10\%]$.

\textbf{Patch Random Drop (PRDrop).} We propose to randomly drop the input patches of self-attention modules to mislead the query-based black-box attacks. The slimming of input patches only slightly decreases the model performance since there is redundant information in inputs, as shown in recent work. Note that the patch drop operation is different from the standard dropout operation where the dropped activations are set to zero. Ours removes part of the patches and keeps the same patches in the rest of the layers. The probability to drop patches controls the trade-off. The probability values we use are in $[1\%, 10\%]$.

\textbf{Patch Attention Perturbation (PAttnPert).} Besides the non-additive randomness in the inputs of self-attention module, we also propose to integrate the non-additive randomness in the Attention of self-attention module. Concretely, we propose to reduce the dimensions of keys and queries by randomly removing some of them. The keys and the queries of patches can still be used to compute the attention since only the dot product between them is required to describe their similarity. In our approach, we propose to randomly remove a certain percentage of dimensions of keys and queries. Such non-additive randomness also demonstrates the high effectiveness against query-based black-box attacks. Similarly, the percentage to remove dimension controls the trade-off.

\section*{D: Combination of Defense Strategies}
\vspace{-0.4cm}
 \begin{figure}[!h]
\centering
    \begin{subfigure}[b]{0.48\textwidth}
    \centering
    \hspace{-0.16cm}\includegraphics[scale=0.65]{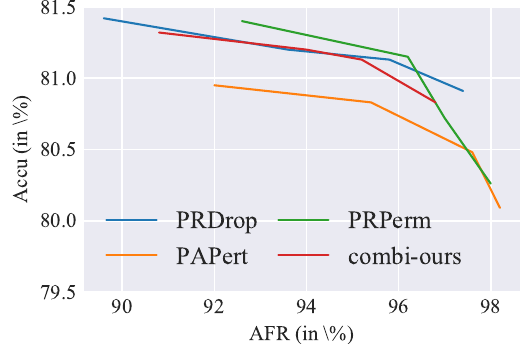}
    \caption{\footnotesize Combination of Our Defense}
    \label{fig:combi_ours}
    \end{subfigure}
    \caption{Combination of Randomness-based Defense on ViTs. We also study the different combinations of the randomness-based defense methods. We show the combinations of our methods, i.e., three different types of randomness-based defense on models. The combination achieves the average defense effect.}
    \label{fig:vit_defense_combi}
\end{figure}

\end{document}